# Apprenticeship Learning using Inverse Reinforcement Learning and Gradient Methods

**Gergely Neu**[*]
Budapest University of Technology and Economics
Műegyetem rkp. 3-9.
H-1111 Budapest, Hungary

**Csaba Szepesvári**[*]
Department of Computing Science
University of Alberta
Edmonton T6G 2E8, AB, Canada

## Abstract

In this paper we propose a novel gradient algorithm to learn a policy from an expert's observed behavior assuming that the expert behaves optimally with respect to some unknown reward function of a Markovian Decision Problem. The algorithm's aim is to find a reward function such that the resulting optimal policy matches well the expert's observed behavior. The main difficulty is that the mapping from the parameters to policies is both nonsmooth and highly redundant. Resorting to subdifferentials solves the first difficulty, while the second one is overcome by computing natural gradients. We tested the proposed method in two artificial domains and found it to be more reliable and efficient than some previous methods.

## 1 INTRODUCTION

The aim of apprenticeship learning is to estimate a policy of an expert based on samples of the expert's behavior. This problem has been studied in the field of robotics for a long time and due to the lack of space we cannot give an overview of the literature. The interested reader might find a short overview in the paper by Abbeel and Ng (2004).

In apprenticeship learning (a.k.a. imitation learning) one can distinguish between *direct* and *indirect approaches*. Direct methods attempt to learn the policy (as a mapping from states, or features describing states to actions) by resorting to a supervised learning method. They do this by optimizing some loss function that measures the deviation between the expert's policy and the policy chosen. The main problem then is that in parts of the state space that the expert tends to avoid the samples are sparse and hence these methods may have difficulties with learning a good policy at such places.

In an indirect method it is assumed that the expert is acting optimally in the environment. In particular, in inverse reinforcement learning the environment is modelled as a Markovian decision problem (MDP) (Ng and Russell, 2000). The dynamics of the environment is assumed to be known (or it could be learnt from samples which might even be unrelated to the samples come from the expert). However, the reward function that the expert is using is unknown. Recently Abbeel and Ng (2004) gave an algorithm which was proven to produce a policy which performs almost as well as the expert, even though it is not guaranteed to recover the expert's reward function (recovering the reward function is an ill-posed problem). This approach might work with less data since it makes use of the knowledge of model of the environment, which can help it in generalizing to the less frequently visited parts of the state space. One problem is that the algorithm of Abbeel and Ng (2004) relies on the precise knowledge of the features describing the reward function, which is not a realistic assumption (for a discussion of this, see Section 6). In particular, we will show that even the correct scales of the features have to be known.

In this paper we propose a gradient algorithm that combines the two approaches by minimizing a loss function that penalizes deviations from the expert's policy like in supervised learning, but the policy is obtained by tuning a reward function and solving the resulting MDP, instead of finding the parameters of a policy. We will demonstrate that this combination can unify the advantages of the two approaches in that it can be both sample efficient and work even when the features are just vaguely known.

---
[*]Computer and Automation Research Institute of the Hungarian Academy of Sciences, Kende u. 13-17, Budapest 1111, Hungary

## 2 BACKGROUND

Let us first introduce some notation: For a subset $S$ of some topological space, $S^\circ$ will be used to denote its interior. For a finite dimensional vector $x$, $\|x\| = \sum_{i=1}^{d} x_i^2$ shall denote its $\ell^2$-norm. Random variables will be denoted by capital letters (e.g., $X, A$), $\mathbb{E}[\cdot]$ stands for expectations.

We assume that the reader is familiar with basic concepts underlying Markovian decision processes (MDPs) (e.g., Puterman 1994), hence we introduce these concepts only to fix the notation. A finite, discounted infinite-horizon total reward MDP is defined by a 5-tuple $M = (\mathcal{X}, \mathcal{A}, \gamma, P, r)$, where

- $\mathcal{X}$ is a finite set of *states*,
- $\mathcal{A}$ is a finite set of *actions*,
- $\gamma \in [0, 1)$ is the *discount factor*,
- $P$ gives the *transition probabilities*; $P(x'|x, a)$ stands for the probability of the transition from state $x$ to $x'$ upon taking action $a$ ($x, x' \in \mathcal{X}, a \in \mathcal{A}$),
- $r$ is the *reward function*; $r : \mathcal{X} \times \mathcal{A} \to \mathbb{R}$; $r(x, a)$ gives the reward incurred when action $a \in \mathcal{A}$ is executed from state $x \in \mathcal{X}$.

A *stationary stochastic policy* (in short: policy) is a mapping $\pi : \mathcal{A} \times \mathcal{X} \to [0, 1]$ satisfying $\sum_{a \in \mathcal{A}} \pi(a|x) = 1, \forall x \in \mathcal{X}$.[1] The value of $\pi(a|x)$ is the probability of taking action $a$ in state $x$. A policy is called deterministic if for any $x$, $\pi(\cdot|x)$ is concentrated on a single action. The class of all stationary stochastic policies will be denoted by $\Pi$.

For a fixed policy, the *value* of a state $x \in \mathcal{X}$ is defined by

$$V^\pi(x) = \mathbb{E}\left[\sum_{t=0}^{\infty} \gamma^t r(X_t, A_t) \middle| X_0 = x\right], \quad (1)$$

where $(X_t, A_t)_{t \geq 0}$ is the sequence of random state-action pairs generated by executing the policy $\pi$. The function $V^\pi : \mathcal{X} \to \mathbb{R}$ is called the *value function* underlying policy $\pi$.

We will also need *action-value functions*. The action-value function, $Q^\pi : \mathcal{X} \times \mathcal{A} \to \mathbb{R}$, underlying policy $\pi$ is defined by

$$Q^\pi(x, a) = \mathbb{E}\left[\sum_{t=0}^{\infty} \gamma^t r(X_t, A_t) \middle| X_0 = x, A_0 = a\right] \quad (2)$$

---
[1] Instead of $\pi(a, x)$ we use $\pi(a|x)$ to emphasize that $\pi(\cdot, x)$ is a probability distribution. Note that in finite MDPs one can always find optimal (stochastic) stationary policies (Puterman, 1994).

with the understanding that for $t > 0$, $A_t \sim \pi(\cdot|X_t)$.

A policy that maximizes the expected total discounted reward over all states is called an *optimal policy*. The *optimal value function* is defined by $V^*(x) = \sup_\pi V^\pi(x)$, while the *optimal action-value function* is defined by $Q^*(x, a) = \sup_\pi Q^\pi(x, a)$.

It turns out that $V^*$ and $Q^*$ satisfy the so-called Bellman optimality equations (e.g., Puterman 1994). In particular,

$$Q^*(x, a) = r(x, a) + \gamma \sum_{y \in \mathcal{X}} P(y|x, a) \max_{b \in \mathcal{A}} Q^*(x, b). \quad (3)$$

We call a policy that satisfies $\sum_{a \in \mathcal{A}} \pi(a|x) Q(x, a) = \max_{a \in \mathcal{A}} Q(x, a)$ at all states $x \in \mathcal{X}$ *greedy* w.r.t. the function $Q$. It is known that all policies that are greedy w.r.t. $Q^*$ are optimal and all stationary optimal policies can be obtained these way.

## 3 APPRENTICESHIP LEARNING

Assume that we observe a sequence of state-action pairs $(X_t, A_t)_{0 \leq t \leq T}$, the 'trace' of some expert. We assume that the expert selects the actions by some unknown policy $\pi_E$: $A_t \sim \pi_E(\cdot|X_t)$. The goal is to recover $\pi_E$ from the observed trace. The simplest solution is of course to use a supervised learning approach: we select a parametric class of policies, $(\pi_\theta)_\theta$, $\pi_\theta \in \Pi$, $\theta \in \mathbb{R}^d$, and try to tune the parameters so as to minimize some loss $J_T(\pi_\theta)$, such as

$$J_T(\pi) = \sum_{x \in \mathcal{X}, a \in \mathcal{A}} \hat{\mu}_T(x)(\pi(a|x) - \hat{\pi}_{E,T}(a|x))^2, \quad (4)$$

where $\hat{\mu}_T(x)$ could be defined by $\hat{\mu}_T(x) = 1/(T+1) \sum_{t=0}^{T} \mathbb{I}_{\{X_t = x\}}$ are the empirical occupation frequencies under the expert's policy and $\hat{\pi}_{E,T}(a|x) = \sum_{t=0}^{T} \mathbb{I}_{\{X_t = x, A_t = a\}} / \sum_{t=0}^{T} \mathbb{I}_{\{X_t = x\}}$ is the empirical estimate of the expert's policy.[2] It is easy to see that $J_T$ approximates the squared loss

$$J(\pi) = \sum_{x \in \mathcal{X}, a \in \mathcal{A}} \mu_E(x)(\pi(a|x) - \pi_E(a|x))^2 \quad (5)$$

uniformly in $\pi$ (the usual concentration results hold for $J_T$, e.g. Györfi et al. (2002)).

The reason $\hat{\pi}_{E,T}$ is not used directly as a 'solution' is that if the state space is large then it will be undefined for a large number of states (where $\hat{\mu}_{E,T}(x) = 0$) with high probability unless the number of samples is enormous.

An alternative to direct policy learning is *inverse reinforcement learning* (Ng and Russell, 2000). The idea is that given the expert's trace, we find a reward function that can be used to explain the performance of the expert. More precisely, the problem is to find a reward function that the behavior of the expert is optimal for. Once the reward function is found, existing algorithms are used to find a behavior that is optimal with respect to it.

---
[2] If a state is not visited by the expert, the policy is defined arbitrarily.

One difficulty in IRL is that solutions are non-unique: e.g. if $r$ is a reward function that recovers the expert's policy then for any $\lambda \geq 0$, $\lambda r$ is also a solution ($r = 0$ is always a solution). For non-trivial problems there are many solutions besides the variants that differ in their scale only.

We propose here to unify the advantages of the direct and indirect approaches by *(i)* taking it seriously that we would like to recover the expert's policy and *(ii)* achieve this through IRL so that we can achieve good generalization at parts of the state space avoided by the expert. We thus propose to find the parameters given a parametric family of rewards $(r_\theta)_\theta \in \Theta$ such that the corresponding (near) optimal policy, $\pi_\theta$, matches the expert's policy $\pi_E$ (more precisely, it's empirical estimate). The proposed method can be written succinctly as the optimization problem

$$J(\pi_\theta) \xrightarrow{\theta} \min! \quad \text{s.t.} \quad \pi_\theta = G(Q_\theta^*), \qquad (6)$$

where $J$ is a loss function (such as (5) or (4)) aimed at measuring the distance of $\pi_E$ and its argument, $Q_\theta^*$ is the optimal action-value function corresponding to the reward function $r_\theta$ and $G$ is a suitable smooth mapping that returns (near) greedy policies with respect to its argument. One possibility, utilized in our experiments, is to use Boltzmann action-selection policies (see (7)).[3]

In this paper we consider gradient methods to solve the above optimization problem. One difficulty with such an approach is that there could be many parameterizations that yield to the same loss. This will be helped with the method of natural gradients, for which the theory is worked out in the next section.

Another difficulty is that the mapping $\theta \mapsto Q_\theta^*$ is non-differentiable. We will, however, show that it is Lipschitz when $r_\theta$ is Lipschitz and hence, by Rademacher's theorem it is differentiable almost everywhere (w.r.t. the Lebesgue measure).

### 3.1 NATURAL GRADIENTS

Our ultimate goal is to find some parameters $\theta$ in a parameter space $\Theta \subset \mathbb{R}^d$ such that the policy $\pi_\theta$ determined by $\theta$ matches the expert's policy $\pi_E$. For facilitating the discussion let us denote the map from the parameter space $\Theta$ to the policy space by $h$ (i.e., $h(\theta) = \pi_\theta$). Thus, our objective function can be written as $\tilde{J}(\theta) = J(h(\theta))$, where $J : \Pi \to \mathbb{R}$ is a (differentiable) objective function defined over $\Pi$ (such as (5)) and the goal is to minimize $\tilde{J}$. Incremental gradient methods implement $\theta_{t+1} = \theta_t - \alpha_t g_t$, where $\alpha_t \geq 0$ is an appropriate step-size sequence and $g_t = g(\theta)$ points in the direction of steepest ascent on the surface $(\theta, \tilde{J}(\theta))_\theta$.

The gradient method with an infinitesimal step-size gives rise to a trajectory $(\theta(t))_{t \geq 0}$. This in turn determines a trajectory $(\pi(t))_{t \geq 0}$ in the policy space, where $\pi(t) = h(\theta(t))$. Since our primary interest is the trajectory in the policy state, it makes sense to determine the gradient direction $g$ in each step such that $\pi(t)$ moves in the steepest descent direction on the surface of $(\pi, J(\pi))_\pi$. We call $g = g(\theta)$ the *natural gradient* if this holds. Amari (1998) gives a method to find the natural gradients using the formalism of Riemannian spaces.

The advantage of this procedure is that the resulting trajectories will be the same for any equivalent parameterization (i.e., if the parameter space is replaced by some other space that is related to the first one through a smooth invertible mapping, with a smooth inverse). In addition, the gradient algorithm that uses natural gradients can be proven to be asymptotically efficient in a probabilistic sense and has the tendency to alleviate the problem of 'plateaus' (Amari, 1998).

In order to define natural gradients we need some definitions. First, we need the generalization of derivatives for mappings $f$ between Banach spaces.[4] The underlying idea is that the gradient (derivative) of $f : U \to V$ provides a linear approximation to the change $f(u + h) - f(u)$:

**Definition 1** (Fréchet derivative). *Let $U, V$ be Banach spaces. $A$ is the* Fréchet-derivative *of $f$ at $u$ if $A : U \to V$ is a bounded linear operator and $\|f(u + h) - f(u) - Ah\|_V = o(\|h\|_U)$. The mapping $f$ then is called* Fréchet differentiable *at $u$.*

In what follows we view $\Pi$ both as a vector space and a complete metric space with some metric $d$. In our application this metirc will be derived from the (unweighted) $\ell^2$-norm, but other choices would also work. The following definition suggests a geometry induced on $\Theta$:

**Definition 2** (Induced metric). *Let $\Theta \subset \mathbb{R}^d$, $\theta \in \Theta^\circ$. We say that $G_\theta \in \mathbb{R}^{d \times d}$ is a pseudo-metric induced by*

---

[3] The benefit of choosing strictly stochastic policies is that if the expert's policy is deterministic, they force the uniqueness of the solution.

[4] A Banach space is a complete normed vector space. In our case it will usually be a Euclidean space, e.g. $\mathbb{R}^d$.

$(h, \Pi, d)$ at $\theta$ if $G_\theta$ is positive semidefinite and

$$d(h(\theta+\Delta), h(\theta)) = \Delta^T G_\theta \Delta + o(\|\Delta\|^2).$$

The essence of this definition is that if the 'distance' between $\theta$ and $\theta+\Delta$ is given by $\Delta^T G_\theta \Delta$ then this distance will match the distance of $h(\theta)$ and $h(\theta+\Delta)$, as $\|\Delta\| \to 0$. It follows from the definition that the induced pseudo-metric is unique.

In the rest of the paper we assume that $\Pi$ is finite dimensional to make the presentation of the results easier. The following proposition is an immediate consequence of the definition of induced pseudo-metrics and the definition of Fréchet differentiability:

**Proposition 1.** *Assume that $h : \Theta \to \Pi$ is Fréchet differentiable at $\theta \in \Theta^\circ$, $\Theta \subset \mathbb{R}^d$, $\Pi = (\Pi, d)$ is a complete, linear metric space. Then $h'(\theta)^T h'(\theta)$ is the pseudo-metric induced by $(h, \Pi, d)$ at $\theta$.*

Natural gradients can be obtained by the following procedure: Let $g(\theta; \varepsilon) = \text{argmax}_{\Delta \in \tilde{S}(\theta,\varepsilon)} \tilde{J}(\theta + \Delta) - \tilde{J}(\theta)$ be the direction of steepest ascent over the 'warped sphere' $\tilde{S}(\theta, \varepsilon) = \{\Delta \in \mathbb{R}^d \mid \|h(\theta+\Delta) - h(\theta)\| = \varepsilon\}$.[5] Then the set of natural gradients is given by

$$\tilde{\nabla}^{(h)} \tilde{J}(\theta) \stackrel{\text{def}}{=} \liminf_{\varepsilon \to 0+} \frac{1}{\varepsilon} g(\theta; \varepsilon).$$

Here the limes inferior of the sets $(g(\theta; \varepsilon))_{\varepsilon>0}$ is meant in the sense of the Painlevé-Kuratowski convergence (Kuratowski, 1966): It then holds that no matter how $\varepsilon$ converges to zero, $g \in \tilde{\nabla}^{(h)} \tilde{J}(\theta)$ defines a direction of steepest ascent on the surface of $J$ at $h(\theta)$.

The following theorem holds:

**Theorem 1.** *Let $J : \Pi \to \mathbb{R}$, $h : \Theta \to \Pi$, $\tilde{J} = J \circ h$. Assume that $J$ is Fréchet differentiable and locally Lipschitz and $h : \Theta \to \Pi$ is Fréchet differentiable at $\theta \in \Theta^\circ$. Let $G_\theta = h'(\theta)^T h'(\theta)$ be the pseudo-metric at $\theta$ induced by $(h, \Pi, d)$. Then $G_\theta^\dagger \nabla \tilde{J}(\theta) \in \tilde{\nabla}^{(h)} \tilde{J}(\theta)$, where $\nabla \tilde{J}(\theta)$ is the ordinary gradient of $\tilde{J}$ at $\theta$ and $G_\theta^\dagger$ denotes the Moore-Penrose generalized inverse of $G_\theta$.*

For the sake of specificity, when it does not cause confusion, we call $G_\theta^\dagger \nabla \tilde{J}(\theta)$ the natural gradient of $\tilde{J}$ at $\theta$. Note that from the construction it follows immediately that the trajectories of $\dot{\theta} = G_\theta^\dagger \nabla \tilde{J}(\theta)$ are covariant for any initial condition.

The proof borrows some ideas from the proof of Theorem 1 in (Amari, 1998). In order to spare some space we only give an outline here: The basic idea is to replace the warped sphere $\tilde{S}(\theta, \varepsilon)$ by the 'sphere' $S_{G_\theta}(\theta, \varepsilon) = \{\Delta \in \mathbb{R}^d \mid \Delta^T G_\theta \Delta = \varepsilon^2\}$. This is justified since the 'sphere' $S_{G_\theta}(\theta, \varepsilon)$ becomes arbitrarily close to $\tilde{S}(\theta, \varepsilon)$ as $\varepsilon \to 0$ and $\tilde{J}$ is sufficiently regular. The next step is to show that for some $C > 0$, $C\varepsilon G_\theta^\dagger \nabla \tilde{J}(\theta)$ is a solution of the optimization problem $\text{argmax}_{\Delta \in S_{G_\theta}(\theta,\varepsilon)} \tilde{J}'(\theta)\Delta$, and this solution tracks closely that of $\text{argmax}_{\Delta \in S_{G_\theta}(\theta,\varepsilon)} \tilde{J}(\theta+\Delta) - \tilde{J}(\theta)$ when $\varepsilon \to 0$.

## 4 CALCULATING THE GRADIENT

In order to calculate the natural gradient we need to calculate the (Fréchet) derivative of $h(\theta) = G(Q_\theta^*)$ and the gradient of $J(h(\theta))$.[6] By the chain rule we obtain $\nabla J(h(\theta)) = J'(h(\theta))h'(\theta)$. Since calculating the derivative of $J$ (or $J_T$) is trivial, we are left with calculating the derivative of $h(\theta)$. As suggested previously, we use a smooth mapping $G$. One specific proposal, that we actually used in the experiments assigns Boltzmann policies to the action-value functions:

$$G(Q)(a|x) = \frac{\exp[\beta Q(x,a)]}{\sum_{b \in \mathcal{A}} \exp[\beta Q(x,b)]}, \qquad (7)$$

where $\beta > 0$ is a parameter that controls how close $G(Q)$ is to a greedy action selection. With this choice

$$\begin{aligned}\frac{\partial \pi_\theta}{\partial \theta_k}(a|x) &= \pi_\theta(a|x) \frac{\partial \ln[\pi_\theta(a|x)]}{\partial \theta_k} \\ &= \pi_\theta(a|x)\beta \left( \frac{\partial Q_\theta^*(x,a)}{\partial \theta_k} - \sum_{b \in \mathcal{A}} \pi_\theta(b|x) \frac{\partial Q_\theta^*(x,b)}{\partial \theta_k} \right).\end{aligned} \qquad (8)$$

Hence, we are left with calculating $\partial Q_\theta^*(x,a)/\partial \theta_k$. We will show that these derivatives can be calculated almost everywhere on $\Theta$ by solving some fixed-point equations similar to the Bellman-optimality equations. For this, we will need the concept of subdifferentials and some basic facts:

**Definition 3** (Fréchet Subdifferentials). *Let $U$ be a Banach space, $U^*$ be its topological dual.[7] The Fréchet subdifferential of $f : U \to \mathbb{R}$ at $u \in U$, denoted by $\partial^- f(u)$ is the set of $u^* \in U^*$ such that*

$$\liminf_{h \to 0, h \neq 0} \|h\|^{-1} [f(u+h) - f(u) - \langle u^*, u \rangle] \geq 0.$$

The following elementary properties follow immediately from the definition (e.g., Kruger 2003):

**Proposition 2.** *Let $(f_i)_{i \in I}$ be a family of real-valued functions defined over $U$ and let $f(u) = \max_{i \in I} f_i(u)$.*

---

[5] Note that $g(\theta; \varepsilon)$ is set-valued.

[6] Remember that $G$ maps action-value functions to policies and $J$ measures deviations to the expert's policy.

[7] When $U = \mathbb{R}^d$ with the $\ell^2$-norm then $U^* = \mathbb{R}^d$ and for $u \in U, v^* \in U^*$, $\langle v^*, v \rangle$ is the normal inner product.

Then if $u^* \in \partial^- f_i(u)$ and $f_i(u) = f(u)$ then $u^* \in \partial^- f(u)$. If $f_1, f_2 : U \to \mathbb{R}$, $\alpha_1, \alpha_2 \geq 0$ then $\alpha_1 \partial^- f_1 + \alpha_2 \partial^- f_2 \subset \partial^- (\alpha_1 f_1 + \alpha_2 f_2)$.

The next result states some conditions under which, in a generalized sense, 'taking a derivative and a limit is interchangeable'. It is extracted from the proof of Proposition 3.4 of Penot (1995):

**Proposition 3.** *Assume that $(f_n)_n$ is a sequence of real-valued functions over $U$ which converge to some function $f$ pointwise. Let $u \in U$, $u_n^* \in \partial^- f_n(u)$ and assume that $(u_n^*)$ is weak$^*$-convergent to $u^*$ and is bounded. Further, assume that the following holds at $u$: For any $\varepsilon > 0$, there exists some index $N > 0$ and a real number $\delta > 0$ such that for any $n \geq N$, $h \in B_U(0, \delta)$,*

$$f_n(u+h) \geq f_n(u) + \langle u_n^*, h \rangle - \varepsilon \|h\|.$$

*Then $u^* \in \partial^- f(u)$.*

Now, we state the main result of this section:

**Proposition 4.** *Assume that the reward function $r_\theta$ is differentiable w.r.t. $\theta$ with uniformly bounded derivatives: $\sup_{(\theta, x, a) \in \mathbb{R}^d \times \mathcal{X} \times \mathcal{A}} \|r'_\theta(x, a)\| < +\infty$. The following statements hold:*

**(1)** *$Q_\theta^*$ is uniformly Lipschitz-continuous as a function of $\theta$ in the sense that for any $(x, a)$ pair, $\theta, \theta' \in \mathbb{R}^d$, $|Q_\theta^*(x, a) - Q_{\theta'}^*(x, a)| \leq L' \|\theta - \theta'\|$ with some $L' > 0$;*

**(2)** *Except on a set of measure zero, the gradient, $\nabla_\theta Q_\theta^*$, is given by the solution of the following fixed-point equation:*

$$\varphi_\theta(x, a) = (r'_\theta(x, a))^T$$
$$+ \gamma \sum_{y \in \mathcal{X}} P(y|x, a) \sum_{b \in \mathcal{A}} \pi(b|y) \varphi_\theta(y, b), \quad (9)$$

*where $\pi$ is any policy that is greedy with respect to $Q_\theta$.*

Note that $(r'_\theta(x, a))^T \in \mathbb{R}^d$. In fact, the above equation can be solved componentwise: The $k$th component of the derivative can be obtained computing the action-value function for the policy $\pi$ using $r'_{\theta, k}$ in place of the reward function.[8]

*Proof.* Let $T : \mathbb{R}^{\mathcal{X} \times \mathcal{A}} \to \mathbb{R}^{\mathcal{X} \times \mathcal{A}}$ be the Bellman operator

$$(TQ)(x, a) = r_\theta(x, a) + \gamma \sum_{y \in \mathcal{X}} P(y|x, a) \max_{b \in \mathcal{A}} Q(y, b).$$

---

[8]Here $r'_{\theta, k}$ is the $k$th component of the derivative of the reward function with respect to $\theta$. We also note in passing that if $r_\theta$ is *convex* in $\theta$ then so is $Q_\theta$. This follows with the reasoning followed in the proof of the first part.

By elementary arguments, if $Q$ is $L$-Lipschitz in $\theta$, then $TQ$ is $R + \gamma L$-Lipschitz in $\theta$, where $R$ is such that for any $\theta, \theta' \in \mathbb{R}^d$, $(x, a) \in \mathcal{X} \times \mathcal{A}$, $|r_\theta(x, a) - r_{\theta'}(x, a)| \leq R\|\theta - \theta'\|$. Choose $Q_0 = 0$. As is well known (e.g., Puterman (1994)), $Q_n = T^n Q_0$ converges to $Q^*$: $Q_\theta^* = \lim_{n \to \infty} T^n Q_0$. Hence, by the previous argument $Q^*$ is $R + \gamma R + \gamma^2 R + \ldots = R/(1-\gamma)$-Lipschitz, proving the first part of the statement.

For the second part, for a policy $\pi$, let us define the operator $S_\pi$, acting over the space of functions $\phi : \mathcal{X} \times \mathcal{A} \to \mathbb{R}^d$, by

$$(S_\pi \phi)(x, a) = (r'_\theta(x, a))^T$$
$$+ \gamma \sum_{y \in \mathcal{X}} P(y|x, a) \sum_{b \in \mathcal{A}} \pi(b|y) \phi(y, b).$$

Let $\pi$ denote a greedy policy w.r.t. $Q_\theta^*$ and let $\pi_n$ be a sequence of policies that are greedy w.r.t. $Q_n$ and where ties are broken so that $\sum_{x \in \mathcal{X}, a \in \mathcal{A}} |\pi(a|x) - \pi_n(a|x)|$ is minimized. It follows that for $n$ large enough, $\pi_n = \pi$. Now, consider the sequence $\varphi_0 = 0$, $\varphi_{n+1} = S_{\pi_n} \varphi_n$. Then for $n$ large enough we have $\varphi_{n+1} = S_\pi \varphi_n$. By induction, $\varphi_n(x, a) \in \partial^-_\theta Q_n(x, a)$ holds for any $n \geq 0$. Indeed, this clearly holds for $n = 0$, while the general case follows by Proposition 2. Now, observe that $S_\pi$ acts separately on each of the $d$ components of its argument and when it is restricted to any of these components, it is a contraction. Hence, $\varphi_n$ converges to the fixed point of $S_\pi$, i.e., the solution of (9). By Proposition 3 the limit is a subdifferential of $\lim_{n \to \infty} Q_n = Q_\theta^*$ (that the condition of this proposition is satisfied follows from the uniform convergence of $\varphi_n$ in $\theta$, which follows since $\|r'_\theta\|$ is uniformly bounded in both $\theta$ and $(x, a)$). Now, since by the first part $Q_\theta^*$ is Lipschitz-continuous in $\theta$, by Rademacher's theorem it is differentiable almost everywhere. It is well-known that if a function is differentiable then its subderivative coincides with its derivative (see e.g. Kruger (2003)). This finishes the proof of the statement. $\square$

## 5 COMPUTER EXPERIMENTS

The goal of the experiments was to assess the efficiency of the algorithm and to test its robustness. We were also interested in how it compares with the algorithm of Abbeel and Ng (2004).

We have implemented three versions of our algorithm: (i) gradient descent using plain gradients, (ii) gradient descent using natural gradients (iii) RPROP using plain gradients.[9] RPROP is a popular adaptive step-size selection algorithm that proved to be very competitive in a number of settings Riedmiller and

---

[9]We tried a "natural RPROP" variant as well (RPROP using natural gradients), but perhaps suprisingly, it give much poorer results than the other algorithms.

Braun (1993). We have implemented the variant described in Igel and Hüsken (2000). We also implemented the "max margin" and the "projection" algorithms described in Abbeel and Ng (2004) to be able to compare the different approaches. Results will be shown for "max margin". The projection algorithm is computationally more efficient, but we have found it less reliable and less data efficient.

We decided to use two test environments: The familiar grid world that has also been used by Abbeel and Ng (2004) and the sailing problem due Vanderbei (1996). The reward function was linear in the unknown parameters.

## 5.1 GRID WORLD

We have run the first series of experiments in grid worlds, where each state is a grid square and the four actions correspond to moves in the four compass directions with 70% success. We constructed the reward function as a linear combination of 5 features ($\phi_i : \mathcal{X} \to \mathbb{R}$, $i = 1, \ldots, 5$), where the features were essentially randomly constructed. The optimal parameter vector $\theta_*$ consists of evenly distributed random values from $[-1, 1]$. In general we try to approximate the reward function with the use of the same set of features that has been used to construct it, but we also examine the situation of unprecisely known features. The size[10] of the grid worlds was set to $10 \times 10$. Value iteration was used for finding the optimal policy (or gradients) in all cases. Unless otherwise stated the data consists of 10 independent trajectories following the optimal policy, each having a length of 100 steps. The learning rate was hand-tuned (with a little effort) and the number of iterations is kept at 100 (usually, convergence happens much earlier). In all cases, the performance measure is the error function $J_E$, defined by (5) and we measure the performance of the optimal policy computed for the found reward function. For the "max margin" algorithm we show the performance of the overall best policy found during the first 100 iterations, thus optimistically biasing these measurements.

We examined the algorithms' behavior when *(i)* the number of the training samples was varied (Figure 1), *(ii)* the features were linearly transformed (Figure 2, Table 1, row 2), and when *(iii)* the features were perturbed (Table 1, row 3).

We see from Figure 1 that for small sample sizes plain gradient is doing the best, while eventually natural gradient becomes the winner. Note that the scale on the $y$ axis is logarithmic, so the differences between these algorithms is not big. "Max margin" also catches up at the end, just like RPROP.

---

[10]Preliminary experiments confirm that our conclusions would not change significantly for other sizes.

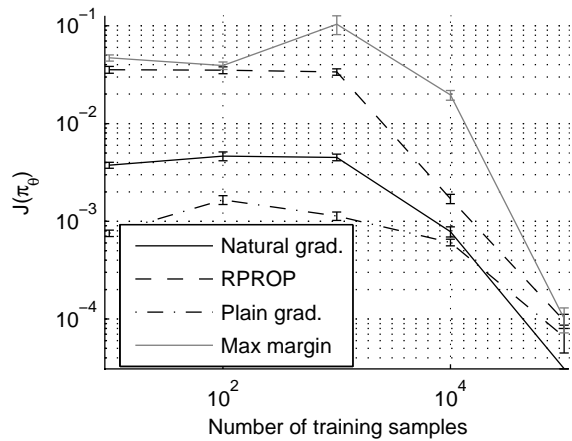

Figure 1: Performance as a function of the number of training samples. Each curve is an average of 10 runs using different samples, with 1/10 s.e. error bars.

Figure 2 shows the effect of transforming the features linearly (the true reward function still remains in the span of the features). Clearly, "Max Marging" suffers badly, while the natural gradient algorithm and RPROP are little affected. Plain gradient descent is slowed down, but eventually converges to good solutions.

In practice, it is not realistic to assume that a subspace containing the reward function is known. To test how the algorithms behave without this assumption we perturbed the features by adding uniform $[-\max(\phi_i)/2, \max(\phi_i)/2]$ random numbers to them. Results are shown in row 3 of Table 1. The results indicate the robustness of natural gradients and RPROP. Both plain gradients and "max margin" suffer large losses under these adverse conditions.

## 5.2 SAILING

We also applied the algorithms to the problem of "sailing" proposed by Vanderbei (1996). In this problem the task is to navigate a boat from one point to another in the shortest possible time. Thus, this is a stochastic shortest path (SSP) problem. Formally, we have a grid of waypoints connected by *legs*, at each waypoint the sailor has to select one of these eight legs to move on to the next waypoint. The state space in this setting is constructed from the actual situation of the boat and the direction from where the wind is blowing at the specific moment. The eight actions of selecting the next waypoint have different costs depending on the direction of the wind: e.g. it costs more time to sail 45 degrees against the wind than to sail 45 degrees in the wind direction etc.. We assume that the wind changes follow a Markov process. The reward function is given using a linear combination of the six features of $(away, down, cross, up, into, delay)$, as defined in Vanderbei (1996) (all defined as a map $\phi : \mathcal{X} \times \mathcal{A} \to \mathbb{R}$). The following weighting was used in the experiments: $\theta_* = (-1, -2, -3, -4, -100000, -3)^T$.

|  | Natural gradients | | RPROP | | Plain gradients | | Max margin | |
|---|---|---|---|---|---|---|---|---|
|  | Mean | Deviation | Mean | Deviation | Mean | Deviation | Mean | Deviation |
| Original | 0.0051 | 0.0010 | 0.0130 | 0.0134 | 0.0011 | 0.0068 | 0.0473 | 0.1476 |
| Transformed | 0 | 0 | 0.0110 | 0.0076 | 0.0256 | 0.0237 | 0.0702 | 0.0228 |
| Perturbed | 0.0163 | 0.0165 | 0.0197 | 0.0179 | 0.1377 | 0.3428 | 0.2473 | 0.3007 |

Table 1: Means and deviations of errors. The row marked 'original' gives results for the original features, the row marked 'transformed' gives results when features are linearly transformed, the row marked 'perturbed' gives results when they are perturbed by some noise.

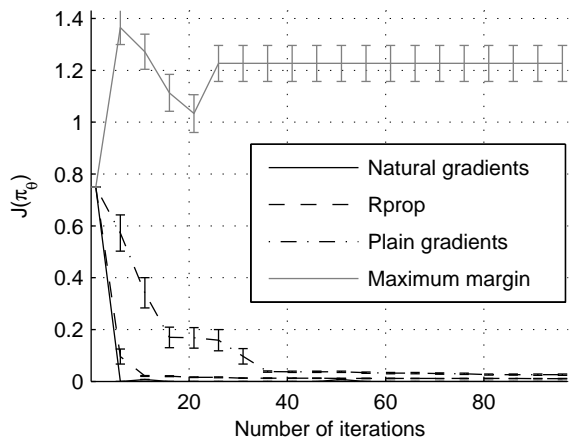

Figure 2: Performance with linearly transformed features. The features were transformed by a (non-singular) square matrix with uniform $[0, 1]$ random elements. Each curve is an average of 25 runs with different scalings of the features, the 1/10 s.e. error bars are also plotted.

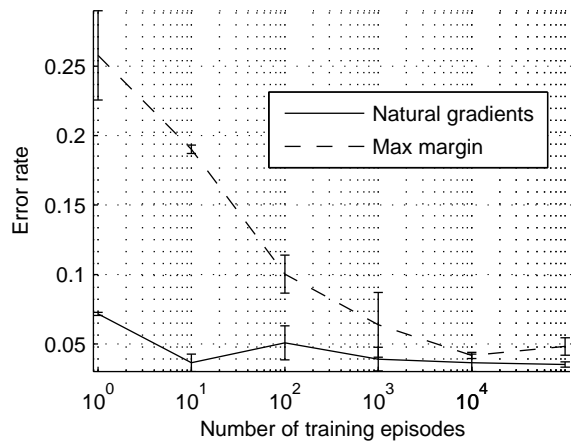

Figure 3: Performance as a function of the number of training episodes. The fraction of states where the found policy differs from the actual optimal policy is plotted against the number of episodes observed., measured by the mean of 5 runs. The 1/2-s.e. error bars are also plotted for both methods.

Results as a function of the number of episodes is shown in Figure 3 for natural gradients and the "max margin" algorithm. In this case the number of iterations is set to 1000 and we again computed the optimal policy with the reward found by the algorithm. As a more tangible performance measure in this case, we show the number of states where the actions selected by the found policy differ from the ones selected by the policy followed by the expert. The results here are shown fro a small lake of size $4 \times 4$.[11] The conclusion is again that the gradient method outperforms the "max margin" algorithm by a significant amount.

## 6 RELATED WORK

Our main concern in this section is the algorithm of Abbeel and Ng (2004). This algorithm returns policies that come with the guarantee that their average total discounted reward computed from the expert's *unknown* reward function is in the $\varepsilon$-vicinity of the expert's performance. We claim that this guarantee will be met only when the scaling of the features in the method and the 'true' scaling match each other. Actually, this observation led us to the algorithms proposed here.

In order to explain why the algorithm of Abbeel and Ng (2004) is sensitive to scalings, we need some background on the algorithm. A crucial assumption in this algorithm is that the reward function is linearly parameterized, i.e., $r(x) = \theta_*^T \phi(x)$, where $\phi : \mathcal{X} \to \mathbb{R}^d$ and $\theta_* \in \mathbb{R}^d$ is the vector of unknown parameters. It follows that the expected total discounted reward is $\theta^T \phi_E$, where $\phi_E \in \mathbb{R}^d$ is the so-called feature expectation underlying the expert. From the trajectory of the expert this can be estimated. In fact, we can define $\phi_\pi$ for any policy $\pi$ and express the expected total discounted reward as $\theta^T \phi_\pi$. The main idea of Abbeel and Ng (2004) is then that it suffices to find a policy $\pi$ whose feature expectations $\phi_\pi$ matches $\phi_E$ since $|\theta_*^T \phi_\pi - \theta_*^T \phi_E| \leq \|\theta_*\|_2 \|\phi_\pi - \phi_E\|_2$.

---

[11]Our preliminary experiments show that the new algorithm performs reasonably for larger problems, too.

However, a major underlying hidden assumption (implicit in the formalism of Abbeel and Ng (2004)) is that the scaling of the features is known. To see this assume that $d = 2$, $\theta_{*,1} = \theta_{*,2} = \sqrt{2}/2$, $\|\phi_\pi - \phi_E\|_2 \leq \varepsilon$ and in particular $\phi_{E,1} = 0$, $\phi_{E,2} > 0$, $\phi_{\pi,1} = -\varepsilon$, $\phi_{\pi,2} = 0$. Further, assume that the features are rescaled by $\lambda = (\lambda_1, \lambda_2)$. In the new scale the expert's performance is $\rho_E(\lambda) = \sqrt{2}/2 \lambda^T \phi_E$ and $\pi$'s performance is $\rho_\pi(\lambda) = \sqrt{2}/2 \lambda^T \phi_\pi = \sqrt{2}/2(\lambda^T \phi_E - \lambda_1 \varepsilon)$. A natural requirement is that for any scaling $\lambda$, $\rho_\pi(\lambda)/\rho_E(\lambda)$ should be lower bound by a positive number (or rather a number close to $1 - \varepsilon$). By straightforward calculations, $\rho_\pi(\lambda)/\rho_E(\lambda) = 1 - (\lambda_1/\lambda_2)\varepsilon/\phi_{E,2} \to -\infty$, hence although $\|\phi_\pi - \phi_E\|_2 \leq \varepsilon$, the actual performance of $\pi$ can be quite far from the performance of the expert if the scaling of the features does not match the scaling used in the algorithm.

More recently, Ratliff et al. (2006) have proposed an algorithm which uses similar ideas to the ones of Abbeel and Ng (2004). Just like Abbeel and Ng (2004) they measure performance with respect to the original reward function and not by the difference of the expert's policy and the policy returned.

## 7 CONCLUSIONS

In the paper we have argued for the advantages of unifying the direct and indirect approaches to apprenticeship learning. The proposed procedure attempts to optimize a cost function, yet it chooses the policy based on a model and thus may overcome problems usually associated with method that directly try to match the expert's policy. Although our method has shown stable behaviour in our experiments, more work is needed to fully explore the limitations of the method. One significant barrier for applying the method (as well as other methods based on IRL) is that it needs to solve MDPs many times. This is problematic since solving an MDP is a challenging problem on its own. One idea is to turn to two time-scale algorithms that run two incremental procedures in parallel, exploiting that a small change to the parameters would likely cause small changes in the solutions; as confirmed by our theoretical results. There are many important directions to continue this work: The present work assumed that states are observed. This could be replaced by the assumption that sufficiently rich features are observed, however, when this is not satisfied the method won't work. For large state-spaces one needs to use function approximation techniques to carry out the computations. It is an open question if the methods would generalize to such settings. Another important direction is to consider infinite MDPs. This presents some technical difficulties, but we expect that the methods could still be generalized to such settings. Yet another interesting direction is to replace the parametric framework with a non-parametric one.

### Acknowledgements

Csaba Szepesvári greatly acknowledges the support received through the Alberta Ingenuity Center for Machine Learning (AICML). This work was supported in part by the PASCAL pump priming project "Sequential Forecasting and Partial Feedback: Applications to Machine Learning". This publication only reflects the authors' views.